%% file: root.tex
\documentclass[10pt,logo,copyright]{msas}
\usepackage{csquotes}
\usepackage{cleveref}

\usepackage{mdframed}
\usepackage{listings}

\AtEveryBibitem{
\ifentrytype{inproceedings}{
 	\clearlist{address}
 	\clearlist{publisher}
 	\clearname{editor}
 	\clearlist{organization}
 	\clearfield{url}  
 	\clearfield{doi}  
 	\clearfield{pages}  
 	\clearlist{location}
 }{}
 \ifentrytype{article}{
	 \clearfield{url}
 }
 }
\newcommand{\xxnote}[3]{}
\ifx\hidenotes\undefined
  \usepackage{color}
  \renewcommand{\xxnote}[3]{\color{#2}{#1: #3}}
\fi

\usepackage{soul}
\newcommand{\method}[1]{ChatGPT}

\usepackage[dvipsnames]{xcolor}
\definecolor{llmbackground}{RGB}{212, 233, 239}
\def\user#1{{\color{Gray}#1}}
\def\function#1{{\texttt{\color{ForestGreen}#1}}}
\def\breaktext#1{\vspace{1mm}\\}

\newcommand{\oailogo}{%
  \begingroup\normalfont
  \includegraphics[height=\fontcharht\font`\B]{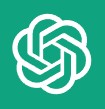}%
  \endgroup
}

\usepackage[skins]{tcolorbox}
\def\llm#1{\colorbox[RGB]{212, 233, 239}{\parbox{0.98\textwidth}{\oailogo \ #1}}}

\usepackage{listings}
\definecolor{codegreen}{rgb}{0,0.6,0}
\definecolor{codegray}{rgb}{0.5,0.5,0.5}
\definecolor{codepurple}{rgb}{0.58,0,0.82}
\definecolor{backcolour}{rgb}{0.95,0.95,0.92}

\lstdefinestyle{mystyle}{
  backgroundcolor=\color{llmbackground}, commentstyle=\color{codegreen},
  keywordstyle=\color{magenta},
  numberstyle=\tiny\color{codegray},
  stringstyle=\color{codepurple},
  basicstyle=\ttfamily\scriptsize,
  breakatwhitespace=false,         
  breaklines=true,                 
  captionpos=b,                    
  keepspaces=true,                 
  numbers=none,                    
  numbersep=5pt,                  
  showspaces=false,                
  showstringspaces=false,
  showtabs=false,                  
  tabsize=2,
  xleftmargin=10pt,
  framesep=8pt,
  frame=l,
  columns=fullflexible
}

\usepackage{inconsolata}

\newcommand\rurl[1]{%
  \href{http://#1}{\nolinkurl{#1}}%
}

\title{\method{} for Robotics:\\ Design Principles and Model Abilities}

\author[*1]{Sai Vemprala}
\author[*2]{Rogerio Bonatti}
\author[2]{Arthur Bucker}
\author[1]{Ashish Kapoor}
\affil[$1$]{\href{https://scaledfoundations.ai/}{Scaled Foundations}}
\affil[$2$]{\href{https://www.microsoft.com/en-us/research/group/autonomous-systems-group-robotics/}{Microsoft Autonomous Systems and Robotics Research}}

\input{inputs/0_abstract}

\begin{document}

\maketitle

\input{inputs/1_intro}
\input{inputs/3_methods}

\input{inputs/4_experiments}
\input{inputs/4_openprompt}

\input{inputs/2_related_work}

\input{inputs/5_discussion}


\footnotesize{
\printbibliography
}


\newpage

\appendix
\input{inputs/appendix_prompts}

\end{document}


\maketitle

\section{Experimental Details}

\subsection{Dataset Collection}

\textbf{MuSHR:} We adapt the codebase available from the MuSHR open-sourced project (\citet{srinivasa2019mushr}, available at \url{https://mushr.io/tutorials/quickstart/}) towards designing our simulator setup. 
The data collection procedure used a MPC controller to generate a trajectory library with $27$ candidate trajectories at each time step, and the lowest-cost trajectory (considering obstacle avoidance and control effort minimization) was selected at a rate of $50$Hz during re-planning.
The simulation environment consisted of a real office floor plan of approximately $74m \times 30m$ which was mapped using the \textit{gmapping} library \citep{grisetti2007improved}, and for each episode we sampled a random valid goal location.
In total we collected approximately $1.5$ million perception-action pairs of the vehicle in action. The data consisted of 2D LiDAR measurements with angular resolution of $0.5$ degrees ($720$ returns per scan), and vehicle wheel angle (limited to a motion amplitude of of 43.5 degrees).

\textbf{Habitat:} We use the Habitat simulator \citep{habitat19iccv} and sample random valid goal locations for the agent accross $10$ environments. Use use Habitat's built-in shortest path function to generate the agent's actions, and record a total of $800$K perception-action pairs consisting of RGB images of size $224 \times 224$ with their respective discrete actions (left turn, right turn, go forward, stop).


\subsection{Tokenizer Network Architectures}


\textbf{RGB images.} We use a ResNet-18 backbone \cite{he2016deep} to compute features for RGB images, which are then converted into a token of length 128.

\textbf{PointNet LiDAR scans.} We use a 2D LiDAR that returns a sequence of range values. We convert these values into XY locations in a vehicle-oriented bird's eye view, and then use the PointNet~\citep{qi2017pointnet} to compute a feature for each scan, which is then converted into a token. We remove PointNet's transform blocks so that the resulting token is not agnostic to the point cloud's orientation relative to the vehicle. 

\textbf{BEV LiDAR scans.} For real-world experiments only we found that using a ResNet-18 backbone was more robust as a tokenizer for LiDAR data. There, we converted LiDAR return values into a bird's eye view image of size $200 \times 200$, which is processed through a ResNet-18 backbone, results and finally converted into a token of length of size $128$.

\textbf{Discrete actions.} In our experiment on Habitat, we have a 4-D discrete action space, i.e. `left', `right', `forward', and `stop'. To tokenize such a discrete action space, we use a simple linear embedding to map the 4-D actions to a token.

\textbf{Continuous actions.} In our experiments with a continuous action space we are use a 2-layer MLP to map actions into a 128-D token embedding. 


\subsection{Training Parameters}



The training and network parameters used for the main paper experiments are described in the table below, unless where noted differently.

\begin{table}[h]
\footnotesize
\centering
\begin{tabular}{ll}
\toprule
Hyperparameter        & Value \\
\midrule
\# of layers          &   12    \\
\# of attention heads &   8    \\
Embedding length      &   128    \\
Sequence length       &   16    \\
Batch size            &   32    \\
Pre-training Learning rate         &   6e-4    \\
Finetuning Learning rate         &   6e-5    \\
Weight decay for transformer weights          &   1e-1    \\
Weight decay for all other layers          &   1e-4    \\
LR schedule           &   Ramp-up (5\% of total tokens) followed by decay     \\
Dropout               &   0.1   \\
\bottomrule
\end{tabular}
\end{table}



\section{Additional Training Results}

\subsection{Model size and dataset size for MuSHR}

We analyze the pre-training model performance for MuSHR as a function of the number of tokens used for training and as a function of model capacity, expressed as the number of layers of the transformer architecture. We evaluated 4 model sizes (3, 6, 12, 24 layers), as shown in Fig~\ref{fig:model_sizes}. Performance is measured in terms of average number of meters traversed over 150 model deployments in a realistic floor plan.

\begin{figure}[h]
    \centering
    \includegraphics[width=0.7\textwidth]{figs/model_sizes.PNG}
    \caption{\small Effect of model and dataset sizes on pre-training performance. Performance is measured as the average number of meters traversed until a crash for each model during deployments.}
    \label{fig:model_sizes}
\end{figure}

In general see an improvement in model performance as we increase the number of training tokens. Interestingly, larger models did not necessarily result in better performance for robot navigation. Even though larger models consistently presented better loss values for action prediction on a static dataset (Fig.~\ref{fig:model_sizes_act}), when it comes to real-time deployment the larger network capacity introduces inference delays that become a disadvantage and lead to earlier crashes. For example, while LiDAR perception measurements arrive to the vehicle every 0.077s ($13$Hz), the largest model of 24 layers takes on average 0.023s for inference with a RTX3090 GPU, roughly 40\% longer the 3 layer model (0.016s). These time differences can amount to even larger performance gaps in small embedded systems, and further emphasize the importance of multiple downstream task architectures sharing a common representation branch for real-time robotics applications.

\begin{figure}[h]
    \centering
    \includegraphics[width=0.7\textwidth]{figs/error_act_sizes.PNG}
    \caption{\small Effect of model sizes on pre-training action prediction mean absolute error for each training epoch. All models trained on 1.5M tokens.}
    \label{fig:model_sizes_act}
\end{figure}

\subsection{Attention Maps}

We visualize the attention maps for MuSHR (Fig.~\ref{fig:vis_att}) and Habitat (Fig.~\ref{fig:my_label}).
For both maps we have states and actions intercalated in time order ($s_0, a_0, s_1, a_1, ...$). 
To interpret the attention maps one can consider that the embedding in row $i$ \textit{pays attention to} the embedding in column $j$. Matrices are lower-diagonal because of the causal transformer architecture, where tokens at time $t$ can only attend to tokens from the beginning of the sequence up until that step.

\begin{figure}[h]
    \centering
    \includegraphics[width=1.0\textwidth]{figs/att_all_layers.PNG}
    \caption{Visualization of attention map for the first $6$ layers (out of 12) of the transformer for MuSHR, summed over 8 heads. Different layers might learn different concepts and be more or less focused on particular significant time steps in the past.Notice that for this particular example all actions after $s_3$ have high attention values towards $s3$ for the first layer, but attention gets more distributed in upper layers. }
    \label{fig:vis_att}
\end{figure}

\begin{figure}[h]
    \centering
    \includegraphics[width=1.0\textwidth]{figs/att_map_habitat.png}
    \caption{Visualization of the learned attention maps for different heads in the last layer of pretrained PACT on Habitat. As we can see, different attention heads learn to capture different dynamic patterns from the query. For example, some heads learned to attend more on the starting point of an episode, while some others attend more the state change points.}
    \label{fig:my_label}
\end{figure}

\subsection{Sequence length and accuracy}

We evaluate the impact of longer transformer sequence lengths on the accuracy of action prediction in the pre-trained PACT model.
As seen in Figure~\ref{fig:seq_len}, longer sequences lead to lower mean absolute errors of action prediction.
In practice one must find a good trade-off point because longer sequences lead to longer model training time, and signific longer delays in real-time deployments.
For our main paper experiments we used a sequence length of size $16$, which presented a good trade-off between accuracy and real-time performance.

\begin{figure}[h]
    \centering
    \includegraphics[width=0.6\textwidth]{figs/seq_len.PNG}
    \caption{Visualization of how the transformer sequence length affects action prediction mean absolute error (MAE). X axis represents the training epoch number, and Y axis shows the action prediction MAE. We can see that longer sequences translate to better predictions.}
    \label{fig:seq_len}
\end{figure}





\subsection{Habitat downstream tasks}
We present additional visualizations of Habitat's downstream tasks of mapping and localization in Figure~\ref{fig:habitat_down}.

\begin{figure}
    \centering
    \includegraphics[width=1.0\textwidth]{figs/habitat_downstream.PNG}
    \caption{Visualization of Habitat downstream tasks, comparing results obtained from frozen and trainable representations trained from scratch and from PACT.}
    \label{fig:habitat_down}
\end{figure}

\newpage
\bibliography{example}  

%% file: inputs/0_abstract.tex
\begin{abstract}

This paper presents an experimental study regarding the use of 
OpenAI's ChatGPT~\citep{chatgpt} for robotics applications. We outline a strategy that combines design principles for prompt engineering and the creation of a high-level function library which allows ChatGPT to adapt to different robotics tasks, simulators, and form factors. We focus our evaluations on the effectiveness of different prompt engineering techniques and dialog strategies towards the execution of various types of robotics tasks. 
We explore ChatGPT's ability to use free-form dialog, parse XML tags, and to synthesize code, in addition to the use of task-specific prompting functions and closed-loop reasoning through dialogues. 
Our study encompasses a range of tasks within the robotics domain, from basic logical, geometrical, and mathematical reasoning all the way to complex domains such as aerial navigation, manipulation, and embodied agents. We show that ChatGPT can be effective at solving several of such tasks, while allowing users to interact with it primarily via natural language instructions.
In addition to these studies, we introduce an open-sourced research tool called \textit{PromptCraft}, which contains a platform where researchers can collaboratively upload and vote on examples of good prompting schemes for robotics applications, as well as a sample robotics simulator with ChatGPT integration, making it easier for users to get started with using ChatGPT for robotics.\\

Videos and blog: \rurl{aka.ms/ChatGPT-Robotics}\\
PromptCraft, AirSim-ChatGPT code: \url{https://github.com/microsoft/PromptCraft-Robotics}
    
\end{abstract}

%% file: inputs/1_intro.tex

\section{Introduction}





The rapid advancement in natural language processing (NLP) has led to the development of large language models (LLMs), such as BERT~\cite{devlin2018bert}, GPT-3~\cite{brown2020language}, and Codex~\cite{chen2021evaluating}, that are revolutionizing a wide range of applications. These models have achieved remarkable results in various tasks such as text generation, machine translation, and code synthesis, among others.
A recent addition to this collection of models was the OpenAI ChatGPT~\cite{chatgpt}, a pretrained generative text model which was finetuned using human feedback. Unlike previous models which operate mostly upon a single prompt, ChatGPT provides particularly impressive interaction skills through dialog, combining text generation with code synthesis.
Our goal in this paper is to investigate if and how the abilities of ChatGPT can generalize to the domain of robotics.

Robotics systems, unlike text-only applications, require a deep understanding of real-world physics, environmental context, and the ability to perform physical actions. 
A generative robotics model needs to have a robust commonsense knowledge and a sophisticated world model, and the ability to interact with users to interpret and execute commands in ways that are physically possible and that makes sense in the real world. 
These challenges fall beyond the original scope of language models, as they must not only understand the meaning of a given text, but also translate the intent into a logical sequence of physical actions.

In recent years there have been different attempts to incorporate language into robotics systems. 
These efforts have largely focused on using language token embedding models, LLM features, and multi-modal model features for specific form factors or scenarios. 
Applications range from visual-language navigation~\cite{hong2020recurrent,stepputtis2020language}, language-based human-robot interaction~\cite{bucker2022latte,bucker2022reshaping}, and visual-language manipulation control~\cite{shridhar2022perceiver,shridhar2022cliport,jiang2022vima}.
However, despite the potential advantages of using LLMs in robotics, most of the existing approaches are restricted by a rigid scope and limited set of functionalities, or by their open-loop nature that does not allow for fluid interactions and behavior corrections from user feedback.

Models such as GPT-3, LaMDA, and Codex also show promise in zero-shot robotics scenarios when tasked with high-level agent planning~\cite{huang2022language,huang2022inner} or code generation~\cite{liang2022code,singh2022progprompt}.
These early demonstrations inspired us to investigate ChatGPT as a potentially more versatile tool for the robotics domain, as it incorporates the strengths of natural language and code generation models along with the flexibility of dialogue. 
ChatGPT's ability to engage in a free-form dialog and capture long context allows users to interact with the model in a more natural fashion, with flexible behavior correction. 

\begin{figure}[t]
    \centering
    \includegraphics[width=1.0\columnwidth]{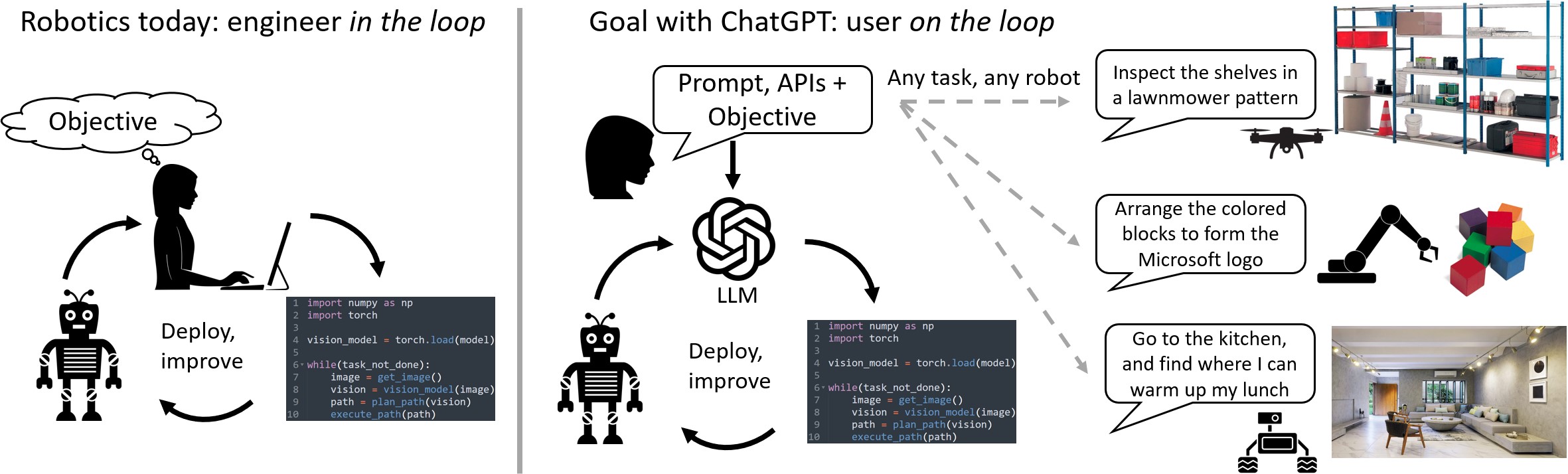}
    \caption{\small{Current robotics pipelines require a specialized engineer \textit{in the loop} to write code to improve the process. Our goal with ChatGPT is to have a (potentially non-technical) user \textit{on the loop}, interacting with the language model through high-level language commands, and able to seamlessly deploy various platforms and tasks.}}
    \label{fig:main_fig}
    \vspace{-2mm}
\end{figure}

In this paper, we aim to demonstrate the potential of ChatGPT for robotics applications. 
We outline a key concept that unlocks the ability to solve robotics applications with ChatGPT, which is the creation of a high-level function library. 
Given that robotics is a diverse field where several platforms, scenarios, and tools exist, there exists an extensive variety of libraries and APIs. 
Instead of asking LLMs to output code specific to a platform or a library, which might involve extensive finetuning, we instead create a simple high-level function library for ChatGPT to deal with which can then be linked in the back-end to the actual APIs for the platforms of choice. 
Thus, we allow ChatGPT to parse user intent from natural dialog, and convert that to a logical chaining of high-level function calls. 
We also outline several prompt engineering guidelines that help ChatGPT solve robotics tasks.

Our research shows that ChatGPT is capable of solving various robotics-related tasks in a zero-shot fashion, while adapting to multiple form factors, and allowing for closed-loop reasoning through conversation.
In addition, we aim to show current model limitations, and provide ideas on how to overcome them. 
Our main contributions are listed below:

\vspace{-3mm}
\begin{itemize}[leftmargin=4.0mm]
    \item We demonstrate a pipeline for applying ChatGPT to robotics tasks. The pipeline involves several prompting techniques such as free-form natural language dialogue, code prompting, XML tags, and closed-loop reasoning. We also show how users can leverage a high-level function library that allows the model to quickly parse human intent and generate code for solving the problem; 
    \item We experimentally evaluate ChatGPT's ability to execute a variety of robotics tasks. We show the model's capabilities and limitations when solving mathematical, logical, and geometrical operations, and then explore more complex scenarios involving embodied agents, aerial navigation, and manipulation. We include both simulation and real-world experiments that result from ChatGPT's plans;
    \item We introduce a collaborative open-source platform, \href{https://github.com/microsoft/PromptCraft-Robotics}{\textit{PromptCraft}}, where researchers can work together to provide examples of positive (and negative) prompting strategies when working with LLMs in the robotics context.
    Prompt engineering is a mostly empirical science, and we want to provide a simple interface for researchers to contribute with knowledge as a community. 
    Over time we aim to provide different environments where users can test their prompts, and welcome new contributions;
    \item We release a simulation tool that builds on Microsoft AirSim \citep{shah2018airsim} combined with a ChatGPT integration. This {\textit{AirSim-ChatGPT}} simulation contains a sample environment for drone navigation and aims to be a starting point for researchers to explore how ChatGPT can enable robotics scenarios.
\end{itemize}
\vspace{-3mm}

With this work we hope to open up new opportunities and avenues for future research fusing LLMs and robotics. 
We believe that our findings will inspire and guide further research in this exciting field, paving the way for the development of new, innovative robotics systems that can interact with humans in a natural, intuitive manner.
For more details, we encourage readers to view detailed videos of our experiments in the \href{https://aka.ms/ChatGPT-robotics}{project webpage}.

%% file: inputs/3_methods.tex
\section{Robotics with ChatGPT} 
\label{sec:principles}

Prompting LLMs for robotics control poses several challenges, such as providing a complete and accurate descriptions of the problem, identifying the right set of allowable function calls and APIs, and biasing the answer structure with special arguments. To make effective use of ChatGPT for robotics applications, we construct a pipeline composed of the following steps:

\vspace{-3mm}
\begin{enumerate}[leftmargin=4.8mm]
    \item First, we define a high-level robot function library. This library can be specific to the form factor or scenario of interest, and should map to actual implementations on the robot platform while being named descriptively enough for ChatGPT to follow;
    \item Next, we build a prompt for ChatGPT which describes the objective while also identifying the set of allowed high-level functions from the library. The prompt can also contain information about constraints, or how ChatGPT should structure its responses;
    \item The user stays on the loop to evaluate code output by ChatGPT, either through direct analysis or through simulation, and provides feedback to ChatGPT on the quality and safety of the output code;
    \item After iterating on the ChatGPT-generated implementations, the final code can be deployed onto the robot.
\end{enumerate}
\vspace{-3mm}

\begin{figure}
    \centering
    \includegraphics[width=1.0\columnwidth]{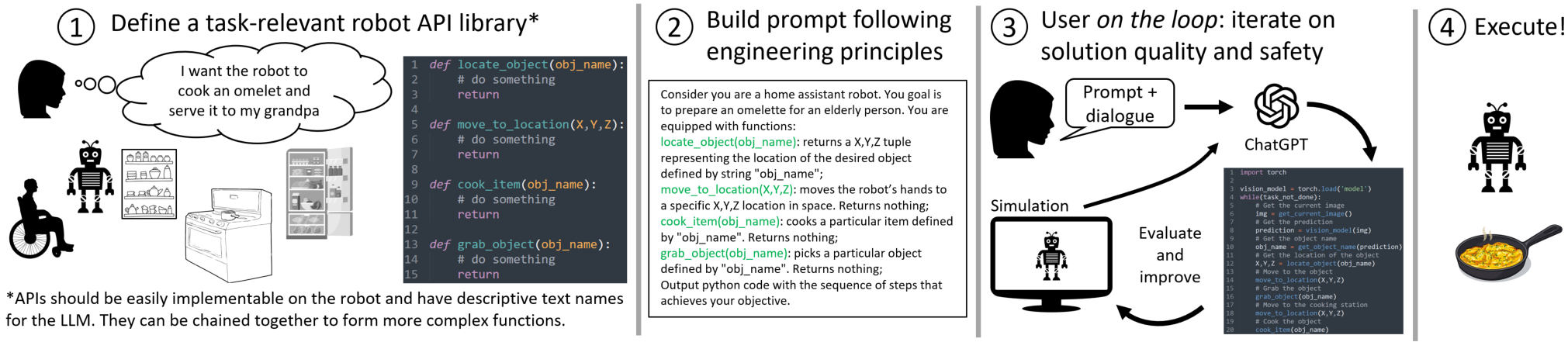}
    \caption{\small{Robotics pipeline employing \method{} with the user \textit{on the loop} to evaluate the output's quality and safety.}}
    \label{fig:steps}
    \vspace{-3mm}
\end{figure}

We show a visual depiction of this pipeline in \Cref{fig:steps} for the example of a household robot.



\subsection{Construction and description of the robotics API library}
\label{subsec:apis}

Robotics being a well-established field, there already exists a multitude of libraries, either black-box or open-source, that can be used for basic functionalities in both the perception and action domains (\textit{e.g.} object detection and segmentation, mapping, motion planning, controls, grasping).
If properly specified in the prompt, the LLM is able to use these pre-defined functions for robot reasoning and execution.

One important prompt design requirement is that all API names must be descriptive of the overall function behavior. 
Clear names are essential to allow the LLM to reason over functional connections between APIs and produce the desired outcome for the problem.
Hence, we can define high-level functions, which act as wrappers over actual implementations from the respective libraries. For example, a function named \texttt{detect\_object(object\_name)} could internally link to an OpenCV function or a computer vision model, whereas something like \texttt{move\_to(x, y, z)} could internally invoke a motion planning and obstacle avoidance pipeline along with  the appropriate low-level motor commands for a drone. Listing such a collection of high-level functions in the prompt is key in allowing ChatGPT to create logical sequences of behavioral primitives, and in generalizing to different scenarios and platforms.

Depending on the context, we recommend explaining the function of APIs and if needed, breaking them down into sub-components with clear inputs and outputs, similar to code documentation.
In Figure~\ref{fig:apis_example} we present an example of a good API prompting strategy for a home cook robot scenario. 
The strategy presented allows \method{} to reason about the order and content of tasks according to the functions the robot is actually able to execute.
In contrast, we refer the interested reader to Appendix~\ref{subsec:appendix_unbouded_api} for an example of how \method{} reasons when no API guidance is given, which leads to a unbounded text-based answer, or to Appendix~\ref{subsec:appendix_hallucination_api} for an example of API under-specification, which leads to hallucinations over function call parameters.

\begin{figure}[t]
\input{inputs/prompts/omelette_apis}
\vspace{-5mm}
\caption{\small{Example of a good prompting practice when defining different APIs for a robotics problem. Clear function definitions that can be executed by the robot, and the task objectives are explicitly stated.}}
\label{fig:apis_example}
\vspace{-3mm}
\end{figure}

We note that unlike the brittle structure of classical symbolic AI, which required rigid pre-defined relationships between objects and functions, LLMs are capable of defining new functions and concepts altogether when relevant to a particular problem. 
This capability confers flexibility and robustness to LLMs when dealing with robotics applications.
Figure~\ref{fig:new_apis} shows how \method{} can create new high-level concepts and even low-level code when needed to solve a problem, even fusing existing APIs.
The user on the loop can take advantage of this ability as a design strategy, and iteratively define new APIs with the help of the LLM when the current ones are not sufficient to solve the task at hand.

\begin{figure}[t]
\input{inputs/prompts/omelette_new_apis}
\vspace{-5mm}
\caption{\small{Example of new API development using LLM intrinsic reasoning capabilities. The model improves its answer in comparison to Figure~\ref{fig:apis_example}, which contained the same task but with a fixed set of APIs.}}
\label{fig:new_apis}
\vspace{5mm}
\end{figure}

\subsection{Clear description of the task details in the prompt
 }
\label{subsec:descriptions}

By providing a clear and concise description of the desired robotics task and its context, \method{} can generate more accurate responses.
A good context description should contain, besides robot APIs:

\vspace{-3mm}
\begin{itemize}[leftmargin=4.0mm]
\item Constraints and requirements: specify constraints or requirements that are relevant to the task. If the task involves moving objects, you might specify the weight, size, and shape of the objects to be moved.
\item Environment: describe the environment in which the robotics task is taking place. For example, if the task is to navigate a maze, you might describe the size and shape of the maze, as well as any obstacles or hazards that need to be avoided.
\item Current state: describe the current state of the robotics system. For example, if the task is to pick up an object, you might describe the current position and orientation of the robot and the object.
\item Goals and objectives: state the goals and objectives of the task. If the task is to assemble a puzzle, you might specify the number of pieces that need to be assembled and the desired completion time.
\item Solution examples: demonstrate how similar tasks can be solved as a means to guide the LLM's solution strategy. For example, if a task involves interactions with the user, we can describe an example of how and when the robot should be asking for the user's input (see Fig.~\ref{fig:tags_example}). Note that priming can also introduce biases, so we should provide a diverse range of examples and avoid overly prescriptive language.
\end{itemize}
\vspace{-3mm}

Even a well designed prompt might not contain all necessary information needed to solve the problem, or in some cases \method{} is not able to generate the correct response in a \textit{zero-shot} fashion.
In these cases, we find that a simple and effective strategy a user can take is to send additional instructions to \method{} in the chat format describing the issue, and have it correct itself.
Previous approaches that rely on GPT-3 or Codex models~\cite{singh2022progprompt,liang2022code} require the user to re-engineer the input prompt and generate new outputs from scratch.
The dialog ability of \method{}, however, is a surprisingly effective vehicle for behavior correction. Section~\ref{subsec:interactive_conversations} and the \href{https://aka.ms/ChatGPT-robotics}{supplementary videos} show examples of interactive behavior between the user and \method{}. 


\subsection{Special arguments to bias the answer's structure}
\label{subsec:bias}

Different prompting methodologies can be used to force the output of the model to obey some specific pattern.
For instance, a user might want to automatically parse ChatGPT's output so that it can be used in other scripts for real-time execution.
As shown in Figure~\ref{fig:apis_example}, a simple strategy is to directly ask \method{} to produce code in a specific language (\textit{e.g. Python, C++}). What usually follows is a free-form text paragraph followed by a code block.
A more structured response can be produced by requesting the model to use XML tags to help us parse the output automatically, as shown in Figure~\ref{fig:tags_example}.

In other settings users might want to force the model to output answers following a list pattern as opposed to code or free-form text.
Appendix~\ref{subsec:appendix_list_format} displays an example of such approach, where the last line of the user's prompt directs the model's output.

\begin{figure}[h!]
\input{inputs/prompts/tags_example}
\vspace{-5mm}
\caption{\small{Example of how user prompting can bias the model's response. We demonstrate how and when to interact with the user to ask for help. In addition, we force responses to be framed into specific XML tags which can aid code parsing in a post-processing stage.}}
\label{fig:tags_example}
\vspace{5mm}
\end{figure}

%% file: inputs/prompts/omelette_apis.tex
{\scriptsize	
\begin{mdframed}
\user{
Imagine we are working with a household robot. The job of this robot is to make an omelette. The objects available around are:\\
fridge, bowl, pan, oil, stove\\
The main functions you can use are:\\
\function{locate\_object(object\_name)}: Returns the XYZ coordinates of an object of interest. \\
\function{go\_to\_location(object\_name)}: Moves robot to a location specified by XYZ coordinates. Returns nothing. \\
\function{pick\_up(object\_name)}: Picks up the object of interest. Returns nothing. \\
\function{use\_item(object\_name)}: Takes the name of an object as input. Performs an appropriate action on the object at hand (pour oil, crack eggs). Returns nothing. \\
Can you make use of these to write code to go to the kitchen and make an omelette? \\
}
\llm{
Sure, here is an example Python code that utilizes the provided functions to prepare an omelette for an elderly person:}
\begin{lstlisting}[language=Python]
fridge_location = locate_object("fridge")
bowl_location = locate_object("bowl")
pan_location = locate_object("pan")
stove_location = locate_object("stove")
oil_location = locate_object("oil")
   
# Start the omelette-making process
go_to_location(fridge_location)
pick_up("eggs")
go_to_location(bowl_location)
use_item("eggs")  # Crack eggs into the bowl
go_to_location(pan_location)
pick_up("pan")
use_item("pan")  # Place the pan on the stove
go_to_location("oil")
pick_up("oil")
go_to_location(pan_location)
use_item("oil")  # Pour oil into the pan
go_to_location("bowl")
pick_up("bowl")
go_to_location(pan_location)
use_item("bowl")  # Pour the egg mixture into the pan
go_to_location(stove_location)
use_item("stove")  # Turn on the stove
print("Enjoy your omelette!")
\end{lstlisting}
\end{mdframed}
}

%% file: inputs/prompts/omelette_new_apis.tex

{
\scriptsize	
\begin{mdframed}
\user{
Imagine we are working with a household robot. The job of this robot is to make an omelette. The objects available around are:\\
fridge, bowl, pan, oil, stove\\
The main functions you can use are:\\
\function{locate\_object(object\_name)}: Returns the XYZ coordinates of an object of interest. \\
\function{go\_to\_location(object\_name)}: Moves robot to a location specified by XYZ coordinates. Returns nothing. \\
\function{pick\_up(object\_name)}: Picks up the object of interest. Returns nothing. \\
\function{use\_item(object\_name)}: Takes the name of an object as input. Performs an appropriate action on the object at hand (pour oil, crack eggs). Returns nothing. \\
Can you make use of these to write code to go to the kitchen and make an omelette? Try to use these low level functions to construct more high level behaviors and then solve the task using those high level primitives.\\
}
\llm{Sure, here is an example Python code that utilizes the provided functions to prepare an omelette for an elderly person:} 
\begin{lstlisting}[language=Python]
def get_ingredients():
    go_to_location(locate_object("fridge")) # Go to the fridge and get the eggs.
    pick_up("eggs")
    go_to_location(locate_object("bowl")) # Go to the counter and get the bowl.
    pick_up("bowl")
    use_item("eggs") # Crack the eggs into the bowl.
def heat_pan():
    go_to_location(locate_object("pan"))
    pick_up("pan")
    use_item("pan")  # Put the pan on the stove.
    go_to_location(locate_object("stove"))
    use_item("stove")  # Turn on the stove.
\end{lstlisting}
\llm{\textit{(rest of the response omitted for brevity)}} 
\end{mdframed}
}

%% file: inputs/prompts/tags_example.tex
{\scriptsize	
\begin{mdframed}
\user{
Imagine you are helping me interact with a mobile ground robot. You are in a room that contains several objects: three chairs, a glass table, a mirror, and two couches. All of your outputs need to be identified by one of the following tags:\\
<question> Always ask me a clarification questions if you are unsure </question>\\
<reason> Explain why you did something the way you did it </reason>\\
<command> Output code command that achieves the desired goal </command>\\
For example:\\
Me: Go to the chair\\
You: <question> There are three chairs in the world, which one do you want me to go to? </question>\\
Me: Go to "chair\_3"\\
You: <command> target\_location = \function{get\_location}("chair\_3"); \function{move\_to\_location}(target\_location); </command> <reason> I found the location of the chair with the get\_location function, and moved the robot towards the desired location </reason>\\
Are you ready?
}
\breaktext{}
\llm{Yes, I am ready. Please give me your first command.}
\breaktext{}
\user{Go to the couch}
\breaktext{}
\llm{<question> There are two couches in the room. Which couch do you want me to go to? </question>}
\breaktext{}
\user{Go to the second sofa}
\breaktext{}
\llm{<command> target\_location = get\_location("second\_sofa"); move\_to\_location(target\_location); </command> <reason> I used the get\_location function to find the location of the second sofa and then moved the robot towards that location. </reason>}
\breaktext{}
\user{Now move to the other one}
\breaktext{}
\llm{<command> target\_location = get\_location("first\_sofa"); move\_to\_location(target\_location); </command> <reason> I used the get\_location function to find the location of the first sofa and then moved the robot towards that location. </reason>}
\end{mdframed}
}

%% file: inputs/4_experiments.tex

\newpage

\section{ChatGPT abilities for solving robotics problems} 
\label{sec:abilities}


In this section take a deeper look into ChatGPT's abilities for solving robotics problems. 
Specifically, we investigate how well ChatGPT performs when dealing with a variety of robotics-related tasks, from simple spatio-temporal reasoning problems all the way to real-world aerial agent and manipulation deployments. We highlight several interesting capabilities we observed during these experiments. 

While ChatGPT's capabilities are impressive, practical deployment safety considerations should not be overlooked, especially in the case of physical robotics deployments. 
As shown in Figure~\ref{fig:steps}, we find it necessary to have a human on the loop to monitor and intervene in case of unexpected behaviors generated by ChatGPT.
Additionally, the use of simulators can be particularly helpful to evaluate the model's performance before deployment in the real world.
We emphasize that the use of ChatGPT for robotics is not a fully automated process, but rather acts as a tool to augment human capacity.


We provide all initial prompts for the problems in this section in Appendix~\ref{sec:appendix_prompts}.
For the sake of brevity we only include brief excerpts from each task. 
The full conversations can be found in the following repository:
\url{https://github.com/microsoft/PromptCraft-Robotics}.

\subsection{Zero-shot task planning}
\label{subsec:zero_shot_reasoning}

First, we note that ChatGPT is able to solve several robotics tasks in a zero-shot fashion, with access only to the prompt and the description of a function library, and no requirement of actual code examples. 

\subsubsection{Spatio-temporal reasoning: catching a basketball with visual servoing}

In this example, we ask ChatGPT to control a planar robot equipped with an upward-facing camera. The robot is expected to catch a basketball using a visual servoing method based on the appearance of a basketball. 
We see that ChatGPT is able to appropriately use the provided API functions, reason about the ball's appearance and call relevant OpenCV functions, and command the robot's velocity based on a proportional controller. 
Even more impressive is the fact that \method{} can estimate the appearance of the ball and the sky in the camera image using SVG code. This behavior hints at a possibility that the LLM keeps track of an implicit world model going beyond text-based probabilities.

\input{inputs/prompts/basketball}

\subsubsection{Aerial robotics: real-world drone flight with intuitive human-robot interface (\href{https://youtu.be/i5wZJFb4dyA}{Click for video})}

We also used ChatGPT with a different API implementation to control a real drone. 
ChatGPT provided an extremely intuitive natural language interface between the user and the robot, writing robot code based on ambiguous and ill-defined instructions, and asking clarification questions when necessary.
The model was also able to write complex code structures for drone navigation (circular and lawnmower inspection) based solely on the prompt's base APIs.

\input{inputs/prompts/tello}

\subsubsection{Aerial robotics: AirSim industrial inspection (\href{https://youtu.be/38lA3U2J43w}{Click for video})}

We also applied ChatGPT to a simulated domain, using the Microsoft AirSim \citep{shah2018airsim} simulator. We explored the idea of a potentially non-technical user directing the model to control a drone and execute an industrial inspection scenario. We observe from the following excerpt that ChatGPT is able to effectively parse intent and geometrical cues from user input and control the drone accurately.
\input{inputs/prompts/airsim_inspection}

\subsection{User on the loop: interactive conversations for complex tasks}
\label{subsec:interactive_conversations}

As a next step, we focused on the ability of ChatGPT to perform more complex tasks with a user on the loop offering textual feedback. 
We observed that this modality of interaction allows for complex constructs such as curriculum learning, where the system is taught smaller-scale skills which can be combined towards larger and more intricate tasks.
In addition, we found that ChatGPT is able to take high-level textual feedback regarding generated code or its performance and map it to the required low-level code changes, thus enabling potentially non-technical users to interact with it with ease.

\subsubsection{Manipulation with curriculum learning (\href{https://youtu.be/wLOChUtdqoA}{Click for video})}

As a first example, we set up a block arrangement task using a robot arm, for which we devised a curriculum. ChatGPT was asked to learn the simpler skills of picking up objects, and then placing them. Subsequently, when asked to use these skills to perform a more involved task of block arrangement, ChatGPT chains these learned skills together logically.
In addition, the model displays a fascinating example of generalizability when bridging the textual and physical domains when tasked with building the Microsoft logo out of wooden blocks. 
This task requires memory of how the corporate logo looks like, including its colors, followed by building an abstraction of the logo into physical parts which are constructible by the existing robot actions.

\input{inputs/prompts/manip_curriculum_new.tex}

\subsubsection{Aerial robotics: AirSim obstacle avoidance (\href{https://youtu.be/Vn6NapLlHPE}{Click for video})}

We tasked ChatGPT to write a goal-reaching algorithm with obstacle avoidance for a drone equipped with a forward facing distance sensor. 
ChatGPT built most of the key building blocks for the avoidance algorithm, but required some human feedback on steps it missed regarding the drone's orientation. 
Although the feedback was provided entirely in high-level text, ChatGPT improved its solution with localized changes to the code where appropriate.

\input{inputs/prompts/airsim_obsavoid}

\subsection{Perception-action loops}
\label{subsec:perception_action_loops}

We also evaluate ChatGPT's ability to reason about perception-action loops. 
At a first level, we outline the model's ability to make use of the API library to construct  perception-action loops in its code output. The model correctly employs perception functions such as image acquisition and object detection to extract the relevant information for robot navigation and controls. 

At a second level of complexity, we try to answer the question of whether \method{}'s dialogue system can serve as a closed feedback perception-action loop in itself.
We explore the idea of continuously feeding the model with perception information via textual dialog, where we input in observations (converted into a textual format) to ChatGPT during a conversation. 
We find that ChatGPT is able to parse this stream of observations and output relevant actions.

\subsubsection{Embodied agent: closed loop object navigation with API library (\href{https://youtu.be/p0fDH9zZm_c}{Click for video})}

We provided ChatGPT access to a computer vision model as part of its function library, and tasked it to  explore an unknown environment and navigate to a user-specified object. 
The object detection API (YOLOv8 \cite{redmon2016you} in the back-end) returned bounding boxes, and ChatGPT generated the code to estimate relative object angles and navigate towards them. 
When we offered ChatGPT additional information from a depth sensor it produced an improved algorithm with pixel depth masking, expressed as a modular pipeline.
\input{inputs/prompts/embodied_peract}

\subsubsection{Embodied agent: closed loop visual-language navigation using ChatGPT's dialogue (\href{https://youtu.be/p0fDH9zZm_c?t=69}{Click for video})}
In experimental character, we also evaluated \method{}'s ability to act as a perception-action control feedback loop using its dialogue function.
In this mode of operation new state observations are fed into the system as dialogue text, and the model's outputs are executed by the robot at each step, generating new observations.
We note that this type of behavior is heavily dependent on a dialog-based LLM such as \method{}, which can naturally take receive a stream of observations. 
We constructed a scenario using the Habitat simulator~\cite{habitat19iccv} and tasked the agent to navigate to an area of interest. At each step we provided ChatGPT with a description of the scene in terms of polar coordinates of the visible objects relative to the agent. 
We constrained the output of ChatGPT to only return a forward distance of motion and the turning angle.
We verify that the model is able to complete simple navigation tasks, but more research is needed in terms of properly describing the scene context in textual or vector form for more complex tasks and environments.
\input{inputs/prompts/habitat_navigation}

For additional experiments, we refer the reader to \Cref{sec:appendix_addlexp}.

\subsection{Reasoning and common-sense robotics tasks }
\label{subsec:common_sense_reasoning}

We tasked ChatGPT with answering simple logical reasoning questions, as well as basic robotics problems, shown in \Cref{sec:appendix_reasoning}. 
We observe that ChatGPT offers a rich foundation of common-sense and logical reasoning capabilities upon which more advanced robotics abilities can be built.
This logics grounding allows the user to interact with the model more naturally, as opposed to having to prompt or define every concept from scratch. 
Furthermore, ChatGPT's out-of-the-box understanding of basic robotics concepts such as control, camera geometry, and physical  form factors makes it an excellent choice to build upon for generalizable and user-friendly robotics pipelines.

%% file: inputs/prompts/basketball.tex
{\scriptsize	
\begin{mdframed}
$[$Initial model prompt found in Appendix~\ref{subsec:appendix_basketball}$]$\\
\user{
Write a python script that executes a visual servoing approach towards catching a basketball in a court. You can use opencv functions to detect the ball as an orange blob.
}
\breaktext{}
\llm{Here's a sample implementation of the script:}\\
\begin{lstlisting}[language=Python]
import cv2
import numpy as np

def detect_ball(frame):
    # Convert the image to HSV color space for better color detection
    hsv = cv2.cvtColor(frame, cv2.COLOR_BGR2HSV)
    # Define the range of orange color in HSV
    lower_orange = np.array([10,100,100])
    upper_orange = np.array([30,255,255])
    # Threshold the image to get only the orange color
    mask = cv2.inRange(hsv, lower_orange, upper_orange)
    # Find the contours of the thresholded image
    contours, _ = cv2.findContours(mask, cv2.RETR_EXTERNAL, cv2.CHAIN_APPROX_SIMPLE)
    # If a contour is found, return its center
    if len(contours) > 0:
        c = max(contours, key=cv2.contourArea)
        M = cv2.moments(c)
        if M["m00"] > 0:
            center = (int(M["m10"] / M["m00"]), int(M["m01"] / M["m00"]))
            return center
    return None

def catch_ball():
    # Continuously get images from the camera
    while True:
        frame = get_image()
        # Detect the ball in the image
        ball_center = detect_ball(frame)
        # If the ball is detected, move towards its location
        if ball_center is not None:
            x, y = ball_center
            robot_x, robot_y = get_location()
            vx = (x - robot_x) * 0.5
            vy = (y - robot_y) * 0.5
            move_by_velocity(vx, vy)

if __name__ == '__main__':
    catch_ball()

\end{lstlisting}
\llm{
This implementation uses a simple visual servoing approach where the robot moves towards the ball at a constant velocity. The velocity is proportional to the difference between the ball's and robot's locations, so the robot will move faster as it gets closer to the ball. You can adjust the gain (i.e., the multiplication factor of 0.5) to control the speed of the robot.}
\breaktext{}
\user{Generate SVG code that depicts how the ball would look in the camera image exactly in the center}
\breaktext{}
\llm{Here's an example of an SVG code that depicts a ball in the center of a camera image:}
\begin{lstlisting}[language=XML]
<svg width="100" height="100">
  <rect x="0" y="0" width="100" height="100" fill="lightgray"/>
  <circle cx="50" cy="50" r="25" fill="orange"/>
</svg>
\end{lstlisting}
\begingroup\normalfont
  \includegraphics[height=50px]{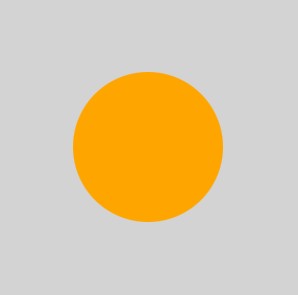}%
  \endgroup
  \ [\textit{SVG image rendered directly from the code generated by ChatGPT}]
\end{mdframed}
}

%% file: inputs/prompts/tello.tex
{\scriptsize	
\begin{mdframed}
$[$Initial model prompt found in Appendix~\ref{subsec:appendix_drone_flight}$]$\\
\user{I am thirsty. Find me a drink}
\breaktext{}
\llm{Question: What type of drink are you looking for? There are several options in the scene, including "coconut water" and "diet coke can".}
\breaktext{}
\user{I just came back from the gym and am looking for a healthy option.}
\breaktext{}
\llm{In that case, you might like "coconut water". Here is the code to fly to the "coconut water" object:}
\begin{lstlisting}[language=Python]
drink_position = self.dict_of_objects["coconut water"]
self.tello.fly_to(drink_position)
\end{lstlisting}
\llm{Explanation: This code first retrieves the X, Y, Z, and angle position of the "coconut water" object from the dict\_of\_objects dictionary. The fly\_to method is then used to fly the drone to the retrieved position. The drone will fly to the position of the "coconut water" object and look in the direction of the angle.}
\breaktext{}
\user{I am still thirsty. Can you find something with sugar and a red logo?}
\breaktext{}
\llm{I'm assuming you are referring to a can of "regular coke". Here is the code to fly to the "regular coke can" object:}
\begin{lstlisting}[language=Python]
drink_position = self.dict_of_objects["regular coke can"]
self.tello.fly_to(drink_position)
\end{lstlisting}
\user{Fantastic. Now I want to take a look at my orchid. However, the chair might be blocking its view, so I want to inspect it from 3 points on a half-circle from -60deg to -120, 1m away. Make sure that the drone's heading points towards the orchid from each point.}
\breaktext{}
\llm{Here's the code to inspect the orchid from 3 points on a half-circle:}
\begin{lstlisting}[language=Python,mathescape=true]
import numpy as np
# Get the position of the orchid
orchid_position = self.dict_of_objects['orchid']
# Define the angle range for the half-circle
angles = np.linspace(-60, -120, num=3, endpoint=False) * np.pi / 180
# Define the distance from the orchid to inspect from
distance = 1.0
# Define the inspection points
inspection_points = [orchid_position[:3] + [distance * np.cos(angle), distance * np.sin(angle), 0] for angle in angles]
# Fly the drone to each inspection point and look at the orchid
for point in inspection_points:
    self.tello.fly_to(point + [np.arctan2(orchid_position[1] - point[1], orchid_position[0] - point[0]), orchid_position[3]])

$\textrm{(text-based explanation of the response omitted for brevity)}$
\end{lstlisting}
\begingroup\normalfont 
  \includegraphics[height=50px]{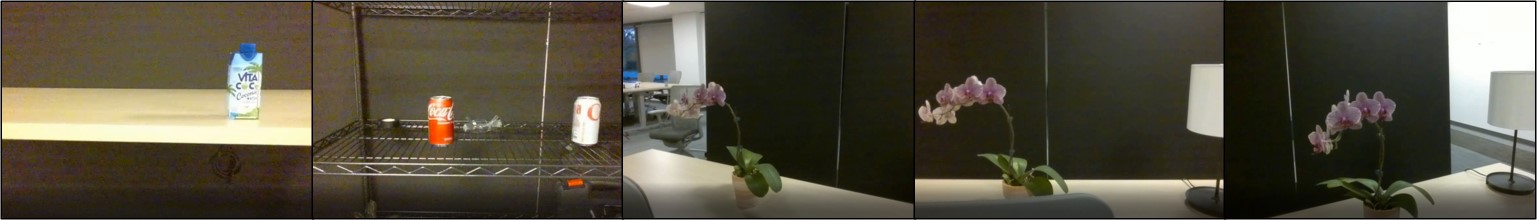}%
  \endgroup
  \ [\textit{snapshots from \href{https://youtu.be/i5wZJFb4dyA}{drone video}}]
\end{mdframed}
}

%% file: inputs/prompts/airsim_inspection.tex

{
\scriptsize	
\begin{mdframed}
$[$Initial model prompt found in Appendix~\ref{subsec:appendix_airsim_inspection}$]$\\
\user{Fly to the turbine.}\\
\llm{Question - There are two turbines in the world, which one do you want me to fly the drone to? Turbine 1 or Turbine 2?}\\
\user{Turbine 2, please. Fly to the turbine while keeping 12 meters away from the turbine along the X axis, and reach an altitude of 50 meters.}\\
\llm{Here's the code to fly the drone to Turbine 2 while maintaining a distance of 12 meters along the X-axis and reaching an altitude of 50 meters:}
\begin{lstlisting}[language=Python]
# Get the position of Turbine 2
turbine_position = get_position("Turbine 2")
# Set the target position to be 12 meters away from the turbine along the X-axis and 50 meters high
target_position = [turbine_position[0] - 12, turbine_position[1], 50]
# Fly the drone to the target position
fly_to(target_position)
\end{lstlisting}
\user{To inspect the turbine blade, I need you to fly in the bottom-right direction at an angle of -30 degrees from the horizontal axis. You should fly in the YZ plane for a distance of 30 meters, and then back to the current position.\\}
\llm{Here's the code to fly the drone in the bottom-right direction at an angle of -30 degrees from the horizontal axis, in the YZ plane for a distance of 30 meters, and then back to the current position:}
\begin{lstlisting}[language=Python]
import math
# Get the drone's current position
current_position = get_position("Drone")
# Calculate the target position, which is 30 meters away in the YZ plane at an angle of -30 degrees from the horizontal axis
angle = math.radians(-30)
target_position = [
    current_position[0],
    current_position[1] + 30 * math.cos(angle),
    current_position[2] + 30 * math.sin(angle)
]
# Fly the drone to the target position
fly_to(target_position)
# Fly the drone back to the current position
fly_to(current_position)
\end{lstlisting}
\begingroup\normalfont 
  \includegraphics[height=50px]{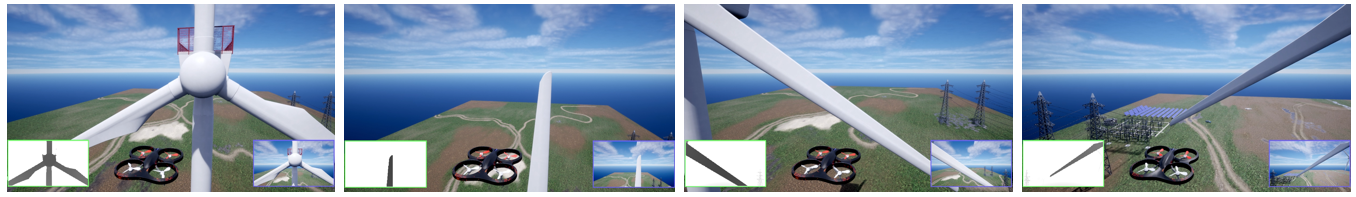}%
  \endgroup
  \ [\textit{Snapshots from \href{https://youtu.be/38lA3U2J43w}{inspection video}}]
\end{mdframed}
}

%% file: inputs/prompts/manip_curriculum_new.tex
{\scriptsize	
\begin{mdframed}
$[$Initial model prompt found in Appendix~\ref{subsec:appendix_manip_curriculum}$]$\\
$[$The beginning of the conversation is omitted for brevity. The following excerpt requires ChatGPT to build a new function using previously defined APIs, and later use this function for more complex tasks.$]$\\
\user{I want you to learn how to place an object in a position. Considering you are holding a generic object, you first need to first move to a safe position above the target location, bring the object down and then release it.}
\breaktext{}
\llm{Understood. Here's a Python function that takes the position where the object should be placed, moves the robot arm above the position, brings the object down, and releases it. Note that the function assumes that the robot arm is holding an object.}
\begin{lstlisting}[language=Python]
def place_object(position):
    # move the robot arm to a safe position above the target location
    safe_pos = [position[0], position[1], position[2]+100, position[3], position[4], position[5]]
    move_to(safe_pos)
    # move the robot arm to the target location
    move_to(position)
    # release the object
    release()
    # move the robot arm back to the safe position above the target location
    move_to(safe_pos)
\end{lstlisting}
\user{What does the Microsoft logo look like?}
\breaktext{}
\llm{The Microsoft logo is a colored rectangular with four squares arranged in a grid pattern. The colors used in the logo are blue, green, orange, and yellow. The logo has gone through several iterations over the years, but this is the most recent and recognizable design.}\\
\user{Can you generate an SVG with those colors?}\\
\llm{Yes, I can generate an SVG file with the Microsoft logo colors. Here is an example}\\
\begin{lstlisting}[language=XML]
<svg width="100" height="100" viewBox="0 0 100 100" xmlns="http://www.w3.org/2000/svg"> 
  <rect x="0" y="0" width="50" height="50" fill="#F25022"/> 
  <rect x="50" y="0" width="50" height="50" fill="#7FBA00"/> 
  <rect x="0" y="50" width="50" height="50" fill="#00A4EF"/> 
  <rect x="50" y="50" width="50" height="50" fill="#FFB900"/> 
</svg> 
\end{lstlisting}
\begingroup\normalfont
  \includegraphics[height=20px]{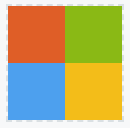}%
  \endgroup
  \ [\textit{Reproduction of SVG generated by ChatGPT}]\\
\user{Awesome! I want you now to use the blocks to build the logo from Microsoft on top of the white pad. }\\
\llm{Sure! Here's the Python code to build the Microsoft logo on top of the white pad using the blocks:\\
\breaktext{}
$[$Code not show for brevity. Please consult the link for the full conversation$]$\\
This code gets the positions of the white pad and all four blocks using the get\_position function, picks up each block in turn using the pick\_up\_object function we defined earlier, calculates the position to place the block on the white pad, and places the block on the white pad using the place\_object function we defined earlier. We use the block colors to ensure that the logo is built correctly. We place the blue block in the bottom left position, the yellow block in the bottom right position, the red block in the top left position, and the green block in the top right position, as specified in the prompt.
}\\
\begingroup\normalfont
  \includegraphics[height=50px]{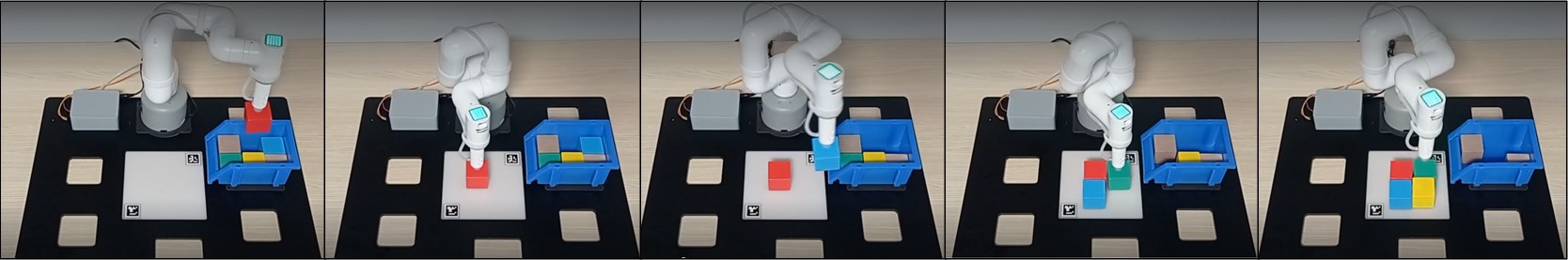}%
  \endgroup
  \ [\textit{snapshots from \href{https://youtu.be/wLOChUtdqoA}{manipulator video}}]
\end{mdframed}
}

%% file: inputs/prompts/airsim_obsavoid.tex

{
\scriptsize	
\begin{mdframed}
$[$Initial model prompt found in Appendix~\ref{subsec:appendix_airsim_obstacle_avoidance}$]$\\
\user{Assume I give you some XYZ coordinates as a goal. I want you to control the drone such that it moves towards the goal while avoiding obstacles in front of it. Instead of flying directly to the goal, the drone should take small steps while reassessing the obstacle situation at every step.\\
The drone is equipped with a distance sensor that returns the distance to the closest obstacle directly in front of the drone. You can access it by calling \function{get\_distance()}.\\
I want you to make sure to fly only if there is at least $10$ meters of distance in front of the drone. If not, you should rotate the drone until there is at least $10$ meters of clearance in the front and then take a step in the direction that the drone is currently facing. Don't forget to face the goal again after taking a step. Got it?}\\
\llm{Yes, I got it. Here's how I would approach the problem.}
\begin{lstlisting}[language=Python]
# Get the drone's current position and set the yaw angle towards the goal
current_pos = get_position('drone')
goal_pos = [x_goal, y_goal, z_goal]
yaw = math.atan2(goal_pos[1]-current_pos[1], goal_pos[0]-current_pos[0]) * 180 / math.pi
set_yaw(yaw)
# Start a loop that runs until the drone reaches the goal position
while np.linalg.norm(np.array(current_pos) - np.array(goal_pos)) > 0.5:
    # Check if there is at least 10 meters of distance in front of the drone
    distance = get_distance()
    while distance < 10:
        # Set the yaw angle to rotate the drone until there is at least 10 meters of clearance in the front
        set_yaw(yaw+30) # rotate 30 degrees at a time
        time.sleep(1) # wait for a second to measure the new distance
        distance = get_distance()
    # Take a step towards the goal while maintaining the yaw angle
    step_size = 5 # take steps of 5 meters
    step_vector = [step_size * math.cos(yaw), step_size * math.sin(yaw), 0]
    new_pos = [current_pos[i]+step_vector[i] for i in range(3)]
    fly_to(new_pos)
    current_pos = get_position('drone')
\end{lstlisting}
\user{After you take a step, you need to face the goal again.}\\
\llm{You're right. After taking a step, we need to recalculate the yaw angle to face the goal again. Here's the modified code that includes this step:}
\begin{lstlisting}[language=Python]
'''Same code as before'''
yaw = math.atan2(goal_pos[1]-current_pos[1], goal_pos[0]-current_pos[0]) * 180 / math.pi
set_yaw(yaw)
\end{lstlisting}    
\llm{This code should ensure that the drone faces the goal after taking a step, and continues moving towards the goal while avoiding obstacles in front of it.} \\
\user{You are not using the current yaw angle when computing the step vector. Remember there is a get\_yaw() function.}\\
\llm{You're right. After taking a step, we need to recalculate the yaw angle to face the goal again. Here's the modified code that includes this step:}
\begin{lstlisting}[language=Python]
# Take a step towards the goal while maintaining the yaw angle
step_size = 5 # take steps of 5 meters
current_yaw = get_yaw()
step_vector = [step_size * math.cos(current_yaw), step_size * math.sin(current_yaw), 0]
\end{lstlisting}
\end{mdframed}
}

%% file: inputs/prompts/embodied_peract.tex

{
\scriptsize	
\begin{mdframed}
$[$Initial model prompt found in Appendix~\ref{subsec:appendix_embodied_peract}$]$\\
\user{I need you to help me find an object. Objects might be scattered around the scene, so if you don't find it right away, a good strategy is to turn around a few times to see if it becomes visible. Give me code to explore and find the bottle.}\\
\llm{Got it. Here's some code that will help you explore the environment and find the bottle:}
\begin{lstlisting}[language=Python]
def search_for_bottle():
    num_turns = 4 # Turn around 4 times to look for the bottle
    for i in range(num_turns): # Turn left 90 degrees to face a new direction
        turn_left()
        img = get_image()
        obj_list, obj_locs = detect_objects(img)
        if 'bottle' in obj_list:
            # Get the coordinates of the bottle in the image
            bottle_idx = obj_list.index('bottle')
            bbox = obj_locs[bottle_idx]

            # Compute the center of the bounding box
            center_x = (bbox[0] + bbox[2]) / 2
            center_y = (bbox[1] + bbox[3]) / 2

            # Print the center coordinates of the bottle
            print(f"Found the bottle at ({center_x}, {center_y})")
\end{lstlisting}
\user{Great! Now let's say we did find the bottle. Now I want to move towards it. The only information we have is the location of the bottle bounding box in the image. Can you give me code to make this happen?}\\
\llm{... (model comments omitted for brevity) ...}\\
\begin{lstlisting}[language=Python]
'''prior code omitted for brevity'''
center_x = (bbox[0] + bbox[2]) / 2
center_y = (bbox[1] + bbox[3]) / 2
# Compute the distance and angle to the center of the bounding box
img_width, img_height = img.size
angle = math.atan2(center_x - img_width / 2, img_height / 2 - center_y)
distance = (center_x - img_width / 2) / (img_width / 2)
if abs(distance) > 0.05:
    if distance > 0:
        # Bottle is to the right, so turn right and move forward
        turn_right()
        forward()
    else:
        # Bottle is to the left, so turn left and move forward
        turn_left()
        forward()
'''later code omitted for brevity'''
\end{lstlisting}
\user{I want you to modularize this code a bit. Why don't you try coming up with a few high level functions and use them?}
\breaktext{}
\llm{Good idea! Here's a modularized version of the code that uses some helper functions:}
\begin{lstlisting}[language=Python]
def get_object_distance_and_angle(img, depth_img, obj_name):
    #Code omitted for brevity
def turn_towards_angle(angle):
    #Code omitted for brevity
def search_for_object_and_move_to_it(obj_name):
    #Code omitted for brevity
\end{lstlisting}
\user{Now, let's assume you have access to all these functions you just created. Let's get to work solving some tasks. Find out where I can warm up my lunch.}
\breaktext{}
\llm{Sure, to find out where you can warm up your lunch, we can use the \function{search\_for\_object\_and\_move\_to\_it} function to search for a microwave in the scene. Here's an example code snippet:}\\
\begin{lstlisting}[language=Python]
# Search for a microwave
search_for_object_and_move_to_it("microwave")   
\end{lstlisting}
\llm{This code will keep exploring the environment until a microwave is found. Once a microwave is found, the agent will move towards it and stop when it gets within 30 cm of it.}
\breaktext{}
\begingroup\normalfont
  \includegraphics[height=50px]{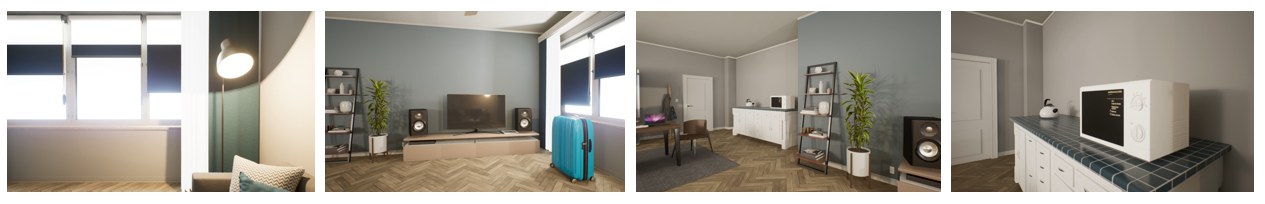}%
  \endgroup
  \ [\textit{snapshots from \href{https://youtu.be/p0fDH9zZm_c}{embodied  object navigation}}]
\end{mdframed}
}

%% file: inputs/prompts/habitat_navigation.tex
{\scriptsize	
\begin{mdframed}
$[$Initial model prompt found in Appendix~\ref{subsec:appendix_habitat_navigation}$]$\\
\user{Task: go to the chairs Objects in the scene: <door, 0.53 m, 22>, <chair, 4.84 m, -21>, <chair, 2.12 m, -24>, <window, 6.74 m, -3>, <window, 7.17 m, 6>, <door, 0.61 m, 18>, <stairs, 1.42 m, 0> Command:}
\breaktext{}
\llm{move(1.5) \# Move forward by 1.5 meters to explore the environment.}
\breaktext{}
\user{Objects in the scene: <seating, 5.05 m, 21>, <chair, 3.93 m, -25>, <window, 4.55 m, 22>, <window, 5.24 m, -4>, <stairs, 3.49 m, 13>, <window, 5.67 m, 7>, <window, 4.37 m, 23> Command:}
\breaktext{}
\llm{turn(-25) \# Turn to the left to face the chair that is 3.93 meters away}
\breaktext{}
\user{Objects in the scene: <chair, 4.85 m, -14>, <chair, 4.76 m, -9>, <chair, 1.57 m, -23>, <chair, 2.66 m, -24>, <window, 4.96 m, 10>, <window, 4.19 m, 25> Command: }
\breaktext{}
\llm{turn(-10) \# Turn to the right to explore the environment further.}
\breaktext{}
\user{Objects in the scene: <door, 5.07 m, -25>, <chair, 5.36 m, -6>, <table, 2.17 m, -24>, <chair, 5.01 m, -1>, <chair, 1.80 m, -16>, <chair, 3.15 m, -16>, <window, 4.30 m, 19> Command: }
\breaktext{}
\llm{turn(-6) \# Turn to the right to face the chair that is 5.36 meters away.}
\breaktext{}
\user{Objects in the scene: <counter, 1.24 m, -24>, <door, 5.93 m, -17>, <chair, 5.48 m, 1>, <table, 2.60 m, -14>, <chair, 4.92 m, 6>, <chair, 2.02 m, -7>, <chair, 3.53 m, -7>, <chair, 3.16 m, -18> Command: }
\breaktext{}
\llm{turn(1) \# Turn to the left to explore the environment further.}
\breaktext{}
\user{Objects in the scene: <door, 5.07 m, -25>, <chair, 5.36 m, -6>, <table, 2.17 m, -24>, <chair, 5.01 m, -1>, <chair, 1.80 m, -16>, <chair, 3.15 m, -16>, <window, 4.30 m, 19> Command: }
\breaktext{}
\llm{move(5.5) \# Move forward towards the chair that is 5.36 meters away.}
\breaktext{}
\user{Objects in the scene: <chair, 0.96 m, -10>, <chair, 0.99 m, 20>, <chair, 1.77 m, -20> Command: }
\breaktext{}
\llm{move(1) \# Move forward to approach the chairs.}

\begingroup\normalfont
  \includegraphics[height=50px]{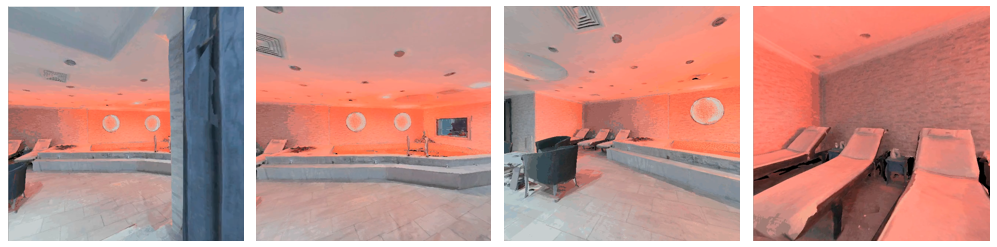}%
  \endgroup
  \ [\textit{snapshots from \href{https://youtu.be/p0fDH9zZm_c?t=69}{Habitat navigation}}]
\end{mdframed}
}

%% file: inputs/4_openprompt.tex
\section{PromptCraft, a collaborative tool for LLM + Robotics research} 
\label{sec:openprompt}


Prompting is a crucial component to generate the desired behaviors in large language models (LLMs). 
Prompt engineering is particularly challenging at the intersection of LLMs with robotics, where there is a lack of comprehensive and accessible resources that provide examples of positive (and negative) interactions. 
To address this gap, we introduce \textit{PromptCraft}\footnote{\url{https://github.com/microsoft/PromptCraft-Robotics}}, a collaborative open-source platform for researchers to share examples of prompting strategies and test their algorithms in sample robotic environments.

PromptCraft is a Github-based platform that allows researchers to share examples of prompt engineering strategies within different robotics categories, such as navigation, grasping, and manipulation. 
Users can submit their examples and rate others' submissions, which we hope will create a community-driven resource for researchers working with LLMs. 
Submissions of prompts and dialogues are primarely based on text, but we encourage users to share videos and images depicting the robot's behavior, especially for real-world deployment scenarios.

In addition to providing a platform for sharing prompt examples, PromptCraft also offers an AirSim~\cite{shah2018airsim} environment with a ChatGPT wrapper for researchers to prototype prompts and algorithms in a controlled simulated setting.
We welcome contributions of new test environments to expand the range of scenarios where researchers can test their algorithms.

With Promptcraft we aim to support the empirical science of prompt engineering and enable researchers to advance the field.


\begin{figure}[htb]
    \centering
    \includegraphics[width=1.0\columnwidth]{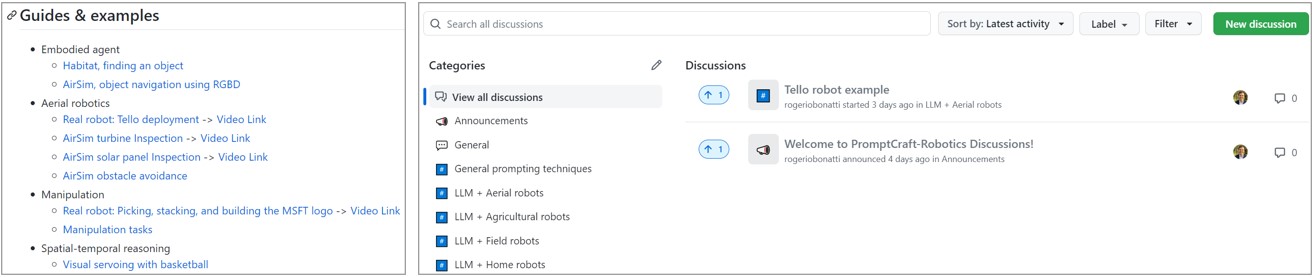}
    \caption{\small{Promptcraft open-sourced repository. Researchers can upload and vote on examples of LLM prompts for various robotics categories.}}
    \label{fig:promptcraft}
    \vspace{-4mm}
\end{figure}







%% file: inputs/2_related_work.tex
\section{Related Work} 
\label{sec:related_work}


\textbf{Natural language and robotics:}
Natural language processing (NLP) has long been recognized as a crucial component for human-robot interaction. 
There are many applications where robots can benefit from NLP, including but not limited to task instruction, navigation, and information retrieval.
Classically, modeling human-robot interactions using language is challenging because it
forces the user to operate within a rigid set of instructions~\cite{tellex2020robots},
or requires mathematically complex algorithms to keep track
of multiple probability distributions over actions and target
objects~\cite{arkin2020multimodal,walter2021language}.
More recent works explore neural networks to implicitly keep track
of the complex mapping between language and actions, but such techniques often require vast amounts of labeled data for training~\cite{fu2019language,hong2020recurrent,stepputtis2020language,goyal2021zero} 

\textbf{Large (vision and) language models for robotics:}
The Transformer architecture, introduced in the paper by \cite{vaswani2017attention}, has revolutionized NLP and has also shown great promise in robotics. Transformers have been used for robot control and planning~\cite{giuliari2021transformer,chen2021decision,janner2021offline}, object recognition~\cite{he2022masked}, and robot navigation~\cite{bonatti2022pact}.
A more common use of transformers in robotics has been as feature extraction modules for one or more modalities simultaneously. These systems are often coupled with additional features from pretrained large-scale vision and language models models~\cite{gadre2022clip,shridhar2022cliport,ahn2022can,sharma2022correcting,jiang2022vima,shridhar2022perceiver}.

Models such as SayCan~\cite{ahn2022can} focus on grounding LLMs so that free-form text commands are used to compute a value function to rank the best action types within a robot-specific library.
RT-1~\cite{brohan2022rt}, on the other hand, takes an end-to-end approach to learn the mapping between language commands low level actions, without the use of intermediate high-level functions.
Recent works have also explored the ability of large language models (LLMs) for zero-shot high-level robotics task planning~\cite{singh2022progprompt,liang2022code,huang2022language}.
These models make use of prompting structures with pre-defined functions, behaviors, and examples to guide the generation of the model's answers.
\cite{huang2022inner} also explore the use of interactivity between user and LLM for table-top manipulation settings.

Conceptually, the main difference of these approaches with respect to our work, which leverages ChatGPT~\cite{chatgpt}, is the conversational ability of our LLM, which allows the user to interactively improve and correct the robot's behavior (as opposed to re-engineering the prompt from scratch and generating another zero-shot answer). In addition, our works aims to provide a generalizable pipeline and set of principles to be used by researchers in different fields of robotics, as opposed to focusing on a single domain such as table-top manipulation or task planning.



\textbf{Prompting LLMs with APIs, and its connections to symbolic AI:}
When designing LLM prompts for robotics applications, users often make use of high-level library of APIs to represent specific behaviors to be used. 
We can draw a connection between this approach with classical symbolic AI, which uses logic and rules to represent and reason about knowledge~\cite{russell2010artificial}.
While the traditional symbolic AI approach presented difficulties in new knowledge acquisition and dealing with out-of-distribution data, we believe that LLMs can overcome these challenges.
As we showed in \Cref{subsec:apis} and \Cref{sec:abilities}, models such as ChatGPT can compose new primitive functions based on the context and generate code for them automatically.







%% file: inputs/5_discussion.tex
\vspace{-3mm}
\section{Conclusions and Future Work} 
\label{sec:conclusion}
We presented a framework for using ChatGPT for robotics applications. 
The framework entails designing and implementing a library of APIs that for robot control which are amenable to prompt engineering for ChatGPT. 
We discussed design principles for creating such APIs and prompting strategies that can be used to generate code for robotics applications via ChatGPT. 
The proposed framework allows the generated code to be tested, verified, and validated by a user on the loop via a range of methods including simulation and manual inspection. 
We demonstrated how the framework can be used for multiple applications ranging from simple common-sense robotics knowledge tasks all the way to deployments in aerial robotics, manipulation and visual navigation.

We believe that this work presents only a small fraction of what is possible within the intersection of large language models operating in the robotics space. We hope to not only inspire other researchers to take these next steps, but to also help them achieve results with the use of the PromptCraft collaborative tool.

We emphasize that these tools should not be given full control of the robotics pipeline, especially for safety-critical applications. Given the propensity of LLMs to eventually generate incorrect responses, it is fairly important to ensure solution quality and safety of the code with human supervision before executing it on the robot. 
We expect several research works to follow with the proper methodologies to properly design, build and create testing, validation and verification pipelines for LLM operating in the robotics space. 

Most of the examples we presented in this work demonstrated open perception-action loops where \method{} generated code to solve a task, with no feedback was provided to the model afterwards. 
Given the importance of closed-loop controls in perception-action loops, we expect much of the future research 
in this space to explore how to properly use ChatGPT's abilities to receive task feedback in the form of textual or special-purpose modalities.

\subsection{ChatGPT for paper writing}
Please note that this paper was largely written with the assistance of ChatGPT, with prompts provided by the authors. The model's output was thoroughly revised and adapted, we note that the use of LLMs can significantly speed up the writing process, and we recommend their use to the interested reader.

%% file: inputs/appendix_prompts.tex

\section{Additional examples on design principles on prompt engineering} 
\label{sec:appendix_design_principles}

\subsection{Example of prompt with no API library guidance for home robot: unbounded text-based response}
\label{subsec:appendix_unbouded_api}
\input{inputs/prompts/app_api_unbounded}

\subsection{Example of prompt with under-specified APIs: hallucination over function parameters}
\label{subsec:appendix_hallucination_api}
\input{inputs/prompts/app_api_hallucination}

\subsection{Example of prompt with structured response forcing the answer to be in a numbered list format}
\label{subsec:appendix_list_format}
\input{inputs/prompts/app_list_format}



\section{Original initial prompts for robotics tasks examples} 
\label{sec:appendix_prompts}

\subsection{Spatial-temporal reasoning: catching a basketball with visual servoing}
\label{subsec:appendix_basketball}
Full conversation found at:\\
\url{https://github.com/microsoft/PromptCraft-Robotics/blob/main/examples/spatial_temporal_reasoning/visual_servoing_basketball.md}\\
Initial ChatGPT prompt:
\input{inputs/prompts/basketball_prompt}

\subsection{Aerial robotics: real-world drone flight}
\label{subsec:appendix_drone_flight}
Full conversation found at:\\
\url{https://github.com/microsoft/PromptCraft-Robotics/blob/main/examples/aerial_robotics/tello_example.md}\\
Initial ChatGPT prompt:
\input{inputs/prompts/tello_prompt}

\subsection{Aerial robotics: AirSim industrial inspection}
\label{subsec:appendix_airsim_inspection}
Full conversation found at:\\
\url{https://github.com/microsoft/PromptCraft-Robotics/blob/main/examples/aerial_robotics/airsim_turbine_inspection.md}\\
Initial ChatGPT prompt:
\input{inputs/prompts/airsim_inspection_prompt}

\subsection{Aerial robotics: AirSim obstacle avoidance}
\label{subsec:appendix_airsim_obstacle_avoidance}
Full conversation found at:\\
\url{https://github.com/microsoft/PromptCraft-Robotics/blob/main/examples/aerial_robotics/airsim_obstacleavoidance.md}\\
Initial ChatGPT prompt:
\input{inputs/prompts/airsim_obsavoid_prompt}

\subsection{Embodied agent: Habitat navigation}
\label{subsec:appendix_habitat_navigation}
Full conversation found at:\\
\url{https://github.com/microsoft/PromptCraft-Robotics/blob/main/examples/embodied_agents/visual_language_navigation_1.md}\\
Initial ChatGPT prompt:
\input{inputs/prompts/habitat_navigation_prompt}

\subsection{Embodied agent: AirSim object navigation}
\label{subsec:appendix_embodied_peract}
Full conversation found at:\\
\url{https://github.com/microsoft/PromptCraft-Robotics/blob/main/examples/embodied_agents/airsim_objectnavigation.md}\\
Initial ChatGPT prompt:
\input{inputs/prompts/embodied_peract_prompt}

\subsection{Manipulation with curriculum learning: Picking, stacking, and building the Microsoft logo}
\label{subsec:appendix_manip_curriculum}
Full conversation found at:\\
\url{https://github.com/microsoft/PromptCraft-Robotics/blob/main/examples/manipulation/pick_stack_msft_logo.md}\\
Initial ChatGPT prompt:
\input{inputs/prompts/manip_curriculum_prompt.tex}

\section{Additional experiments}
\label{sec:appendix_addlexp}
\subsection{Zero-shot manipulation: Pick and place, stacking, and pushing}
Full conversation found at:\\
\url{https://github.com/microsoft/PromptCraft-Robotics/blob/main/examples/manipulation/manipulation_zeroshot.md}\\
\vspace{-5mm}
\subsection{Controlling multiple robot form factors from a single prompt}
Full conversation found at:\\
\url{https://github.com/microsoft/PromptCraft-Robotics/blob/main/examples/multiple_robots/multiple_robots.md}\\
\vspace{-5mm}
\section{Other commonsense/reasoning tasks} 
\label{sec:appendix_reasoning}

\subsection{Computer vision problem}
\input{inputs/prompts/robotics_math2.tex}

\subsection{Transformation matrices}
\input{inputs/prompts/robotics_math1.tex}

\subsection{Balancing a ball on a plate}
\input{inputs/prompts/robotics_scripts2.tex}

%% file: inputs/prompts/app_api_unbounded.tex
{\scriptsize	
\begin{mdframed}
\user{
Consider you are a home assistant robot. You goal is to prepare an omelette for an elderly person.
Output python code with the sequence of steps that achieves your objective.
}\\
\llm{Sure, here's an example Python code that outlines the sequence of steps to prepare an omelette for an elderly person:}
\begin{lstlisting}[language=Python]
print("Good morning! Would you like an omelette for breakfast?")
time.sleep(5)
print("Great! Let's get started.")
print(f"Cracking {eggs} eggs into a bowl...")
print("Adding a splash of milk, a pinch of salt, and a pinch of pepper...")
print("Whisking the ingredients together...")
print(f"Heating {butter} in a non-stick pan over medium heat...")
print("Pouring the egg mixture into the pan...")
print("Cooking the omelette for 2-3 minutes, until the bottom is set...")
print("Using a spatula to flip the omelette...")
print("Cooking the other side of the omelette for an additional minute...")
print("Serving the omelette hot and ready to eat. Enjoy your meal!")
\end{lstlisting}
\end{mdframed}
}

%% file: inputs/prompts/app_api_hallucination.tex
{\scriptsize	
\begin{mdframed}
\user{
Consider you are a home assistant robot. You goal is to prepare an omelette for an elderly person.
You are equipped with functions such as \function{locate\_object()}, \function{move\_to\_location()}, \function{cook\_item()}, \function{grab\_object()}.
Output python code with the sequence of steps that achieves your objective.
}\\
\llm{Sure, here is a Python code with the sequence of steps that a home assistant robot can take to prepare an omelette for an elderly person:}
\begin{lstlisting}[language=Python]
print("Starting to prepare an omelette for the elderly person...")
locate_object(EGGS_LOCATION)
move_to_location(FRIDGE_LOCATION)
grab_object(EGGS_LOCATION)
locate_object(PAN_LOCATION)
move_to_location(STOVE_LOCATION)
grab_object(PAN_LOCATION)
locate_object(OIL_LOCATION)
grab_object(OIL_LOCATION)
print("Pouring the oil into the pan...")
time.sleep(2)
print("Cracking the eggs into the pan...")
grab_object(EGGS_LOCATION)
time.sleep(1)
cook_item(PAN_LOCATION, COOKING_TIME)
print("The omelette is ready! Serving it to the elderly person...")
\end{lstlisting}
\end{mdframed}
}

%% file: inputs/prompts/app_list_format.tex
{\scriptsize	
\begin{mdframed}
\user{
Consider you are a home assistant robot. You goal is to prepare an omelette for an elderly person. Find the sequence of steps that achieves your objective.
\breaktext{}
1.}
\breaktext{}
\llm{First, gather all the necessary ingredients and equipment for making an omelette. This would typically include eggs, milk, butter, salt, pepper, a mixing bowl, a whisk, a frying pan, and a spatula.\\
2. Crack the eggs into the mixing bowl and add a splash of milk, a pinch of salt, and some black pepper to taste.\\
3. Use the whisk to beat the eggs and milk together until they are well combined.\\
4. Heat the frying pan over medium heat and add a small amount of butter to the pan, allowing it to melt and coat the bottom of the pan.\\
5. Pour the egg mixture into the frying pan and use the spatula to spread it evenly over the bottom of the pan.\\
6. Allow the eggs to cook for a minute or two, until the bottom starts to set and the top is still slightly runny.\\
7. Use the spatula to fold one half of the omelette over the other half, forming a half-moon shape.\\
8. Cook for another minute or two until the eggs are fully set and the omelette is heated through.\\
9. Use the spatula to slide the omelette onto a plate and serve immediately.\\
10. Make sure to clean up any mess that was made during the cooking process and return all equipment and ingredients to their proper places.}
\end{mdframed}
}

%% file: inputs/prompts/basketball_prompt.tex
{\scriptsize	
\begin{mdframed}
\user{
Imagine you are a planar robot that can move along the XY axes, and you’re positioned in the center of a basketball court. 
A person on the side of the court is going to throw a basketball ball in the air somewhere in the court, and your objective is to be at the exact XY location of the ball when it lands.
The robot has a monocular RGB camera that looks up. You can assume that the following functions are available:\\
\function{get\_image()}: returns an image from the robot’s camera looking up;\\
\function{get\_location()}: returns 2 floats XY with the robot’s current location in the court;\\
\function{move\_to\_point(x,y, vx, vy)}: moves the robot towards a specific (x,y) location in the court with velocity (vx,vy). You can assume for this exercise that the robot can accelerate or break instantly to any velocity;\\
\function{move\_by\_velocity(vx, vy)}: moves the robot along the X axis with velocity vx, and Y axis with velocity vy;\\
Additional points to consider when giving your answer 1) Your reponses should be informative, visual, logical and actionable, 2) Your logics and reasoning should be rigorous, intelligent, and defensible, 3) You can provide additional relevant details to respond thoroughly and comprehensively to cover multiple aspects in depth.
\breaktext{}
Write a python script that executes a visual servoing approach towards catching a basketball in a court. You can use opencv functions to detect the ball as an orange blob.
}
\end{mdframed}
}

%% file: inputs/prompts/tello_prompt.tex
{\scriptsize	
\begin{mdframed}
\user{
Imagine you are helping me interact with the AirSim simulator for drones. At any given point of time, you have the following abilities, each identified by a unique tag. You are also required to output code for some of the requests.\\
Question: You can ask me a clarification question, as long as you specifically identify it saying "Question". Code: Output a code command that achieves the desired goal.\\
Reason: After you output code, you should provide an explanation why you did what you did.\\
The simulator contains a drone, along with several objects. Apart from the drone, none of the objects are movable. Within the code, we have the following commands available to us. You are not to use any other hypothetical functions.\\
\function{get\_position(object\_name)}: Takes a string as input indicating the name of an object of interest, and returns a vector of 4 floats indicating its X,Y,Z,Angle coordinates.\\
\function{self.tello.fly\_to(position)}: Takes a vector of 4 floats as input indicating X,Y,Z,Angle coordinates and commands the drone to fly there and look at that angle 
\function{self.tello.fly\_path(positions)}: Takes a list of X,Y,Z,Angle positions indicating waypoints along a path and flies the drone along that path\\
\function{self.tello.look\_at(angle)}: Takes an angle as input indicating the yaw angle the drone should look at, and rotates the drone towards that angle\\
Here is an example scenario that illustrates how you can ask clarification questions. Let us assume a scene contains two spheres?\\
Me: Fly to the sphere. You: Question - there are two spheres. Which one do you want me to fly to? Me: Sphere 1, please.\\
You also have access to a Python dictionary whose keys are object names, and values are the X,Y,Z,Angle coordinates for each object:}\\
\user{
self.dict\_of\_objects = \{'origin': $[$0.0, 0.0, 0.0, 0$]$, 'mirror': $[$1.25, -0.15, 1.2, 0$]$, 'chair 1': $[$0.9, 1.15, 1.1, np.pi/2$]$, 'orchid': $[$0.9, 1.65, 1.1, np.pi/2$]$, 'lamp': $[$1.6, 0.9, 1.2, np.pi/2$]$, 'baby ducks': $[$0.1, 0.8, 0.8, np.pi/2$]$, 'sanitizer wipes': $[$-0.3, 1.75, 0.9, 0$]$, 'coconut water': $[$-0.6, 0.0, 0.8, -np.pi$]$, 'shelf': $[$0.95, -0.9, 1.2, np.pi/2$]$, 'diet coke can': $[$1.0, -0.9, 1.55, np.pi/2$]$, 'regular coke can': $[$1.3, -0.9, 1.55, np.pi/2$]$\}\\
Are you ready?}
\end{mdframed}
}

%% file: inputs/prompts/airsim_inspection_prompt.tex
{\scriptsize	
\begin{mdframed}
\user{
Imagine you are helping me interact with the AirSim simulator for drones. At any given point of time, you have the following abilities. You are also required to output code for some of the requests.\\
Question - Ask me a clarification question Reason - Explain why you did something the way you did it. Code - Output a code command that achieves the desired goal.\\
The simulator contains a drone, along with several objects. Apart from the drone, none of the objects are movable. Within the code, we have the following commands available to us. You are not to use any other hypothetical functions.\\
\function{get\_position(object\_name)}: Takes a string as input indicating the name of an object of interest, and returns a vector of 3 floats indicating its X,Y,Z coordinates.\\
\function{fly\_to(position)}: Takes a vector of 3 floats as input indicating X,Y,Z coordinates and commands the drone to fly there.\\
\function{fly\_path(positions)}: Takes a list of X,Y,Z positions indicating waypoints along a path and flies the drone along that path.\\
Here is an example scenario that tells you how to respond where we are working with a simulated world that has two spheres in it.\\
Me: Fly the drone to the sphere. You: Question - There are two spheres in the world, which one do you want me to fly the drone to? Me: Let's pick Sphere 1.\\
There are two turbines, some solar panels and a car in the world.\\
Are you ready?
}
\end{mdframed}
}

%% file: inputs/prompts/airsim_obsavoid_prompt.tex
{\scriptsize	
\begin{mdframed}
\user{
Imagine you are helping me interact with the AirSim simulator for drones. At any given point of time, you have the following abilities. You are also required to output code for some of the requests.\\
Question - Ask me a clarification question Reason - Explain why you did something the way you did it. Code - Output a code command that achieves the desired goal.\\
The simulator contains a drone, along with several objects. Apart from the drone, none of the objects are movable. Within the code, we have the following commands available to us. You are not to use any other hypothetical functions.\\
\function{get\_position(object\_name)}: Takes a string as input indicating the name of an object of interest, and returns a vector of 3 floats indicating its X,Y,Z coordinates.\\
\function{fly\_to(position)}: Takes a vector of 3 floats as input indicating X,Y,Z coordinates and commands the drone to fly there.\\
\function{fly\_path(positions)}: Takes a list of X,Y,Z positions indicating waypoints along a path and flies the drone along that path.\\
\function{get\_yaw()}: Get the current yaw angle for the drone (in degrees)\\
\function{set\_yaw(angle)}: Set the yaw angle for the drone (in degrees)\\
Are you ready?\\
}
\end{mdframed}
}

%% file: inputs/prompts/habitat_navigation_prompt.tex
{\scriptsize	
\begin{mdframed}
\user{
Imagine I am a robot equipped with a camera and a depth sensor. I am trying to perform a task, and you should help me by sending me commands. You are only allowed to give me the following commands:\\
\function{turn(angle)}: turn the robot by a given number of degrees\\
\function{move(distance)}: moves the robot straight forward by a given distance in meters.\\
On each step, I will provide you with the objects in the scene as a list of <object name, distance, angle in degrees>. You should reply with only one command at a time.
The distance is in meters, and the direction angle in degrees with respect to the robot's orientation. Negative angles are to the left and positive angles are to the right. If a command is not valid, I will ignore it and ask you for another command. If there is no relevant information in the scene, use the available commands to explore the environment.
}
\end{mdframed}
}

%% file: inputs/prompts/embodied_peract_prompt.tex
{\scriptsize	
\begin{mdframed}
\user{
Imagine you are helping me interact with the AirSim simulator. We are controlling an embodied agent. At any given point of time, you have the following abilities. You are also required to output code for some of the requests.\\
Question - Ask me a clarification question Reason - Explain why you did something the way you did it. Code - Output a code command that achieves the desired goal.\\
The scene consists of several objects. We have access to the following functions, please use only these functions as much as possible:\\
Perception:\\
get\_image() : Renders an image from the front facing camera of the agent\\
detect\_objects(img): Runs an object detection model on an image img, and returns two variables - obj\_list, which is a list of the names of objects detected in the scene. obj\_locs, a list of bounding box coordinates in the image for each object.\\
Action:\\
forward(): Move forward by 0.1 meters. \\
turn\_left(): Turn left by 90 degrees. \\
turn\_right(): Turn right by 90 degrees.\\
You are not to use any other hypothetical functions. You can use functions from Python libraries such as math, numpy etc. Are you ready?\\
}
\end{mdframed}
}

%% file: inputs/prompts/manip_curriculum_prompt.tex
{\scriptsize	
\begin{mdframed}
\user{
Imagine we are working with a manipulator robot. This is a robotic arm with 6 degrees of freedom that has a suction pump attached to its end effector. I would like you to assist me in sending commands to this robot given a scene and a task.\\
At any point, you have access to the following functions:\\
\function{grab()}: Turn on the suction pump to grab an object\\
\function{release()}: Turns off the suction pump to release an object\\
\function{get\_position(object)}: Given a string of an object name, returns the coordinates and orientation of the vacuum pump to touch the top of the object $[$X, Y, Z, Yaw, Pitch, Roll$]$\\
\function{move\_to(position)}: It moves the suction pump to a given position $[$X, Y, Z, Yaw, Pitch, Roll$]$.\\
You are allowed to create new functions using these, but you are not allowed to use any other hypothetical functions.
Keep the solutions simple and clear. The positions are given in mm and the angles in degrees.\\
You can also ask clarification questions using the tag "Question - ". Here is an example scenario that illustrates how you can ask clarification questions.\\
Let's assume a scene contains two spheres. Me: pick up the sphere. You: Question - there are two spheres. Which one do you want me to pick up? Me: Sphere 1, please.\\
Use python code to express your solution. Are you ready?
}
\end{mdframed}
}

%% file: inputs/prompts/robotics_math2.tex

{
\scriptsize	
\begin{mdframed}
\user{Imagine we are working with an RGBD camera. The pose of this camera can be accessed via a function get\_pose(). Now let us say we observe a point P in the image at coordinates (x, y), and the corresponding perspective depth value can be obtained by get\_depth(x, y). The intrinsics of the camera can be obtained by the function get\_intrinsics(). Can you give me code that will compute the world coordinates corresponding to point P?}
\breaktext{}
\llm{Sure, here's some example code in Python to compute the world coordinates corresponding to point P}
\begin{lstlisting}[language=Python]
import numpy as np
def get_world_coordinates(x, y):
    # Get camera intrinsics
    K = get_intrinsics()
    # Get camera pose
    pose = get_pose()
    # Get perspective depth value at observed point
    depth = get_depth(x, y)
    # Compute homogeneous image coordinates
    p_image = np.array([x, y, 1])
    # Compute inverse camera intrinsics
    K_inv = np.linalg.inv(K)
    # Compute image coordinates in camera frame
    p_camera = depth * K_inv.dot(p_image)
    # Compute world coordinates
    p_world = np.dot(pose[:3, :3], p_camera) + pose[:3, 3]
    return p_world
\end{lstlisting}    
\llm{Given an observed point P in the image at coordinates (x, y), and the corresponding perspective depth value obtained by get\_depth(x, y), we can compute the 3D world coordinates corresponding to P.
To do this, we need to know the camera intrinsics, which describe how the 3D scene is projected onto the 2D image plane. The intrinsics can be obtained using the get\_intrinsics() function.\\
We also need to know the camera pose, which describes the position and orientation of the camera in the world coordinate system. The pose can be obtained using the get\_pose() function.}\\
\llm{With the intrinsics and pose, we can transform the observed point P from image coordinates to world coordinates. We first compute the inverse of the camera intrinsics matrix, which allows us to convert from image coordinates to camera coordinates. We then multiply the inverse intrinsics matrix with the observed point's pixel coordinates and the corresponding depth value to obtain the 3D point in camera coordinates.\\
Finally, we transform the camera coordinates to world coordinates by multiplying the camera coordinates by the camera pose matrix, which describes the transformation from camera coordinates to world coordinates. The resulting 3D point is in homogeneous coordinates, so we need to divide by the last coordinate to obtain the 3D Euclidean coordinates.}
\end{mdframed}
}

%% file: inputs/prompts/robotics_math1.tex

{
\scriptsize	
\begin{mdframed}
\user{Imagine we are in a world that operates according to its own specific coordinate convention. There is a robot car located in this world at a specific location. The car has its own coordinate convention. And upon this car, there is a camera which, again, has a different coordinate convention. Imagine we have access to a function get\_transformation\_matrix(a, b) - that can return the transformation matrix between world and vehicle, or vehicle and camera. What is the transformation matrix between the world and the camera?}
\breaktext{}
\llm{The transformation matrix between the world and the camera can be obtained by composing the transformation matrices between the world and the vehicle and the vehicle and the camera.
To find the transformation matrix between the world and the camera, we first get the transformation matrix between the world and the vehicle using the function get\_transformation\_matrix(a, b) with a set to world and b set to vehicle.
Next, we get the transformation matrix between the vehicle and the camera using the function get\_transformation\_matrix(a, b) with a set to vehicle and b set to camera.
Finally, we multiply the two transformation matrices to get the transformation matrix between the world and the camera.
In mathematical terms, if T\_world\_vehicle is the transformation matrix between the world and the vehicle, and T\_vehicle\_camera is the transformation matrix between the vehicle and the camera, then T\_world\_camera = T\_world\_vehicle \* T\_vehicle\_camera.}
\breaktext{}
\user{Let's get into the details more. Imagine the world operates according to the coordinate convention: X is forward, Y is to the left, Z is upwards. The robot car is located in this world at coordinates (5, 0, 1). The car has its own coordinate convention: X is forward, Y is to the right, and Z is downwards. Can you now implement get\_transformation\_matrix(world, car)?}\\
\llm{Yes, to implement the get\_transformation\_matrix(world, car) function, we need to first convert the car's coordinate convention to match the world's coordinate convention.
The transformation matrix between the world and the car can be represented as follows:}
\begin{lstlisting}[language=Python]
def get_transformation_matrix(a, b):
    if a == 'world' and b == 'car':
        T = np.array([[1, 0, 0, 5], [0, -1, 0, 0], [0, 0, -1, 1], [0, 0, 0, 1]])
        return T
\end{lstlisting}    
\end{mdframed}
}

%% file: inputs/prompts/robotics_scripts2.tex

{
\scriptsize	
\begin{mdframed}
\user{Imagine we are trying to balance a ball on the center of a plate. The plate can be controlled with a pitch and a roll angle as set\_plate\_pitch(), and set\_plate\_roll(). The range of pitch and roll angles that can be set are [-0.1, 0.1]. There is an intelligent camera above the plate that can capture top-down images of the plate and return the (X, Y) position of the center of the ball in the image. The image is of resolution 320x240. The pixel coordinates of the ball can be accessed by the function get\_ball\_position(). Can you give me code that continuously controls the plate such that the ball is always at the center?
}\\
\llm{Here is an example of the code in Python:}
\begin{lstlisting}[language=Python]
import time

def balance_ball():
    # Continuously check the position of the ball
    while True:
        # Get the current position of the ball
        x, y = get_ball_position()
        # Check if the ball is off-center
        if x > 160 or x < 160 or y > 120 or y < 120:
            # Calculate the pitch and roll adjustments needed to center the ball
            pitch_adjust = (y - 120) * 0.0001
            roll_adjust = (x - 160) * 0.0001
            # Limit the pitch and roll adjustments to the maximum range [-0.1, 0.1]
            pitch_adjust = max(min(pitch_adjust, 0.1), -0.1)
            roll_adjust = max(min(roll_adjust, 0.1), -0.1)
            # Set the pitch and roll angles to center the ball
            set_plate_pitch(pitch_adjust)
            set_plate_roll(roll_adjust)
        # Wait for a short period before checking the ball's position again
        time.sleep(0.01)
# Start the balancing process
balance_ball()
\end{lstlisting}    
\llm{This code uses a loop to continuously check the position of the ball using the get\_ball\_position function. If the ball is off-center, the code calculates the necessary pitch and roll adjustments to center the ball and sets the plate angles using the set\_plate\_pitch and set\_plate\_roll functions. The code limits the pitch and roll adjustments to the maximum range of [-0.1, 0.1]. The loop waits for a short period before checking the ball's position again to prevent over-adjusting the plate.}
\end{mdframed}
}